%% file: main.tex
\documentclass[sigconf]{acmart}

\AtBeginDocument{%
  \providecommand\BibTeX{{%
    Bib\TeX}}}

\copyrightyear{2026}
\acmYear{2026}
\setcopyright{cc}
\setcctype{by}
\acmConference[WWW '26] {Proceedings of the ACM Web Conference 2026}{April 13--17, 2026}{Dubai, United Arab Emirates.}
\acmBooktitle{Proceedings of the ACM Web Conference 2026 (WWW '26), April 13--17, 2026, Dubai, United Arab Emirates}
\acmISBN{979-8-4007-2307-0/2026/04}
\acmPrice{}
\acmDOI{10.1145/3774904.3792461}

\acmSubmissionID{rfp2391}

\usepackage{amsmath,amsfonts}
\usepackage{algorithmic}
\usepackage{array}
\usepackage{textcomp}
\usepackage{stfloats}
\usepackage{url}
\usepackage{verbatim}
\usepackage{graphicx}

\usepackage{subcaption}

\usepackage{nicematrix}
\usepackage{algorithm}

\usepackage{amssymb}
\usepackage{xcolor}
\usepackage{booktabs}
\usepackage{multirow}
\usepackage{adjustbox}
\usepackage{hyperref}
\usepackage{enumitem}
\usepackage{cellspace}
\newcommand{\fst}[1]{\color{red}{\underline{\textbf{#1}}}}
\newcommand{\snd}[1]{\color{blue}{\underline{#1}}}

\def \methodname {TRACE}
\def \diffusionname {SPDM}

\newcommand{\TwoRow}[1]{\multirow{2}{*}{#1}}

\def\BibTeX{{\rm B\kern-.05em{\sc i\kern-.025em b}\kern-.08em
    T\kern-.1667em\lower.7ex\hbox{E}\kern-.125emX}}
\usepackage{balance}

\begin{document}

\title{TRACE: Trajectory Recovery with State Propagation Diffusion for Urban Mobility}

\author{Jinming Wang}
\orcid{0009-0000-3314-9758}
\affiliation{%
  \institution{University of Exeter}
  \department{Department of Computer Science}
  \streetaddress{Stocker Rd}
  \city{Exeter}
  \country{United Kingdom}}
\email{jw1294@exeter.ac.uk}

\author{Hai Wang}
\orcid{0000-0001-7317-507X}
\affiliation{%
  \institution{Jingdong Logistics}
  \streetaddress{No. 18 Kechuang 11 Street}
  \city{Beijing}
  \country{China}}
\email{hai@seu.edu.cn}

\author{Hongkai Wen}
\orcid{0000-0003-1159-090X}
\affiliation{%
  \institution{University of Warwick}
  \department{Department of Computer Science}
  \streetaddress{Kirby Corner Road}
  \city{Coventry}
  \country{United Kingdom}
}
\email{hongkai.wen@warwick.ac.uk}

\author{Geyong Min}
\orcid{0000-0003-1395-7314}
\affiliation{%
 \institution{University of Exeter}
  \department{Department of Computer Science}
  \streetaddress{Stocker Rd}
  \city{Exeter}
  \country{United Kingdom}}
\email{G.Min@exeter.ac.uk}

\author{Man Luo*} \thanks{*Corresponding author.}
\orcid{0000-0002-7346-9024}
\affiliation{%
  \institution{University of Exeter}
  \department{Department of Computer Science}
  \streetaddress{Stocker Rd}
  \city{Exeter}
  \country{United Kingdom}}
\email{M.Luo@exeter.ac.uk}

\renewcommand{\shortauthors}{Jinming Wang, Hai Wang, Hongkai Wen, Geyong Min, and Man Luo}

\input{0_Abstract}


\input{0_Abstract}
\maketitle


\input{1_Introduction}

\input{2_Preliminaries}

\input{3_Methodology}

\input{4_Experiment}

\input{5_RelatedWorks}

\input{6_Conclusion}




\balance
\bibliographystyle{ACM-Reference-Format}
\bibliography{ref}

\end{document}

%% file: 0_Abstract.tex
\begin{abstract}
High-quality GPS trajectories are essential for location-based web services and smart city applications, including navigation, ride-sharing and delivery. However, due to low sampling rates and limited infrastructure coverage during data collection, real-world trajectories are often sparse and feature unevenly distributed location points. Recovering these trajectories into dense and continuous forms is essential but challenging, given their complex and irregular spatio-temporal patterns. In this paper, we introduce a novel diffusion model for \textbf{TRA}jectory r\textbf{EC}overy named \textbf{TRACE}, which reconstruct dense and continuous trajectories from sparse and incomplete inputs. At the core of TRACE, we propose a State Propagation Diffusion Model (SPDM), which integrates a novel memory mechanism, so that during the denoising process, TRACE can retain and leverage intermediate results from previous steps to effectively reconstruct those hard-to-recover trajectory segments. Extensive experiments on multiple real-world datasets show that TRACE outperforms the state-of-the-art, offering $>$26\% accuracy improvement without significant inference overhead. Our work strengthens the foundation for mobile and web-connected location services, advancing the quality and fairness of data-driven urban applications. Code is available at:~\url{https://github.com/JinmingWang/TRACE}
\end{abstract}

%% file: 1_Introduction.tex
\input{__Figure_Intro}

\section{Introduction}

The proliferation of GPS-enabled devices has woven a digital fabric over our urban landscapes, generating vast streams of trajectory data. Location-based web services and smart city applications rely on high-quality trajectories - for navigation, ride-hailing, last-mile delivery, and traffic management. The promise of these technologies - optimized routes, efficient traffic management, and equitable service access - hinges on the availability of continuous, high-quality trajectory data. In the wild, however, trajectories recorded by mobile and edge devices are often \textit{sparse} and \textit{irregular}: sampling rates may drop to save energy, infrastructure coverage is uneven across regions, and device storage constraints limit logging. As a result, real-world trajectories are frequently sampled at low rates, yielding spatially skewed location points with large, uneven time gaps. As shown in Fig.~\ref{fig:intro}, such fragmented paths can misrepresent actual movement, degrading downstream task accuracy and fairness.

To bridge this gap, trajectory recovery serves as an essential pre-processing step, aiming to reconstruct dense and uniform trajectories from sparse and incomplete observations. This task is naturally \textit{conditional generation}: given i) a sparse observation and ii) a set of query timestamps, infer the missing locations, so as to obtain a complete and dense trajectory. While various methods have been developed, they face significant limitations. Map-matching techniques~\cite{MapMatching,MTrajRec} are constrained by the availability of detailed digital road networks, which is not always accessible everywhere, while sequence models like RNNs~\cite{RNN,DHTR} often suffer from error accumulation over long, unobserved segments. Transformer-based architectures~\cite{Transformer,AttnMove} show promise by capturing long-range dependencies, but still struggle when spatio-temporal distribution of input locations is highly skewed. Generative models like VAEs~\cite{VAE,TrajVAE,TrajBert} and GANs~\cite{GAN,TrajGANContinuous} have framed recovery as a conditional generation task, thus improving fidelity of the generated trajectories, but VAEs can over-smooth or suffer posterior collapse, GANs risk mode collapse and unstable - and neither explicitly aggregates evidence across long, irregular gaps.

Recently, diffusion models~\cite{DDPM,StableDiffusion} have emerged as a dominant force in generative modeling, demonstrating remarkable success in related tasks including trajectory generation~\cite{DiffTraj,diff_rn_traj,ControlTraj} and recovery of time-series~\cite{PriSTI}. 

Applying them naively to trajectory recovery, however, overlooks domain-specific irregularities: unlike typical time-series data, observations (i.e. of agent locations) arrive at uneven, often long time gaps, are spatially constrained by road topology and agent kinematics, and the query timestamps to be filled rarely align with the observed ones. In standard diffusion, each denoising step treats the input almost in isolation: a UNet-style denoising network~\cite{UNet} repeatedly re-encodes the same sparse evidence and static conditions (timestamps, masks, coarse priors) without carrying forward the intermediate features, i.e. what was already inferred. Early steps - dominated by low SNR noise - tend to wash out weak cues from distant observations; later steps then lack a consolidated memory to resolve long, ambiguous gaps, leading to over-smoothed trajectory segments, temporal jitter, or hallucinated detours — especially under fast sampling schedules. 

To overcome these challenges, we introduce \methodname{}, a novel diffusion-based trajectory recovery framework specifically engineered for the irregular spatio-temporal structure of trajectory data. At the core of \methodname{} is the State Propagation Diffusion Model (\diffusionname{}), which fundamentally redesigns the denoising processes to be \textit{memoryful} rather than step-wise isolated. Given sparse observations and a set of query timestamps, \methodname{} first builds a unified conditioning context that fuses observation/query masks, time gaps, and lightweight priors (e.g., linear interpolation for easy segments; optional context such as day-of-week or user/route IDs). The proposed \diffusionname{} then performs conditional denoising while carrying a compact, multi-scale hidden state across diffusion steps, i.e., its stateful denoising network acts as a memory mechanism, allowing multi-scale spatio-temporal features to be retained, accumulated, and reused across successive denoising steps. This propagated state summarizes geometry and motion cues discovered early and feeds them forward, turning a sequence of isolated denoising steps into a coherent, memoryful refinement process. Concretely, in \methodname{} a UNet backbone processes the conditioning context, while a step-aware state propagation module updates the hidden state to concentrate capacity on long, uncertain gaps and avoid re-processing those stable regions. This not only avoids redundant computation but also provides richer, evolving context, i.e., knowledge from previous denoising steps can now serve as auxiliary conditions for subsequent steps, empowering the model to reconstruct challenging, irregular segments with far greater accuracy. The technical contributions of this paper are summarized as follows: 

\begin{itemize}[leftmargin=6mm]
\item To the best of our knowledge, we are the first to formalize trajectory recovery as conditional generation under severe spatial/temporal irregularity, and propose a novel diffusion-based framework, \methodname{}, that is explicitly designed to handle the sparse and irregular nature of real-world trajectory data. This provides a more reliable data foundation for a broad spectrum of downstream applications, including location-based web services.
\item We introduce the State Propagation Diffusion Model (\diffusionname{}), a novel architecture that augments the standard diffusion models with a step-aware state propagation mechanism. \diffusionname{} carries compact, multi-scale states across the denoising process -  by reusing this intermediate knowledge as conditioning, \diffusionname{} avoids redundant processing but concentrates capacity on the hard-to-recover segments while skipping the already stable ones.
\item We evaluate \methodname{} across multiple real-world datasets and sparsity regimes. Extensive experiments show that \methodname{} consistently outperforms strong baselines by significant margins. Ablations confirm that the proposed \diffusionname{} mechanism is the key, achieving up to a 26.65\% improvement in recovery accuracy over the standard diffusion model, establishing a new state-of-the-art without significant inference overhead.

\end{itemize}

%% file: __Figure_Intro.tex
\begin{figure}
    \centering
    \includegraphics[width=0.9\linewidth]{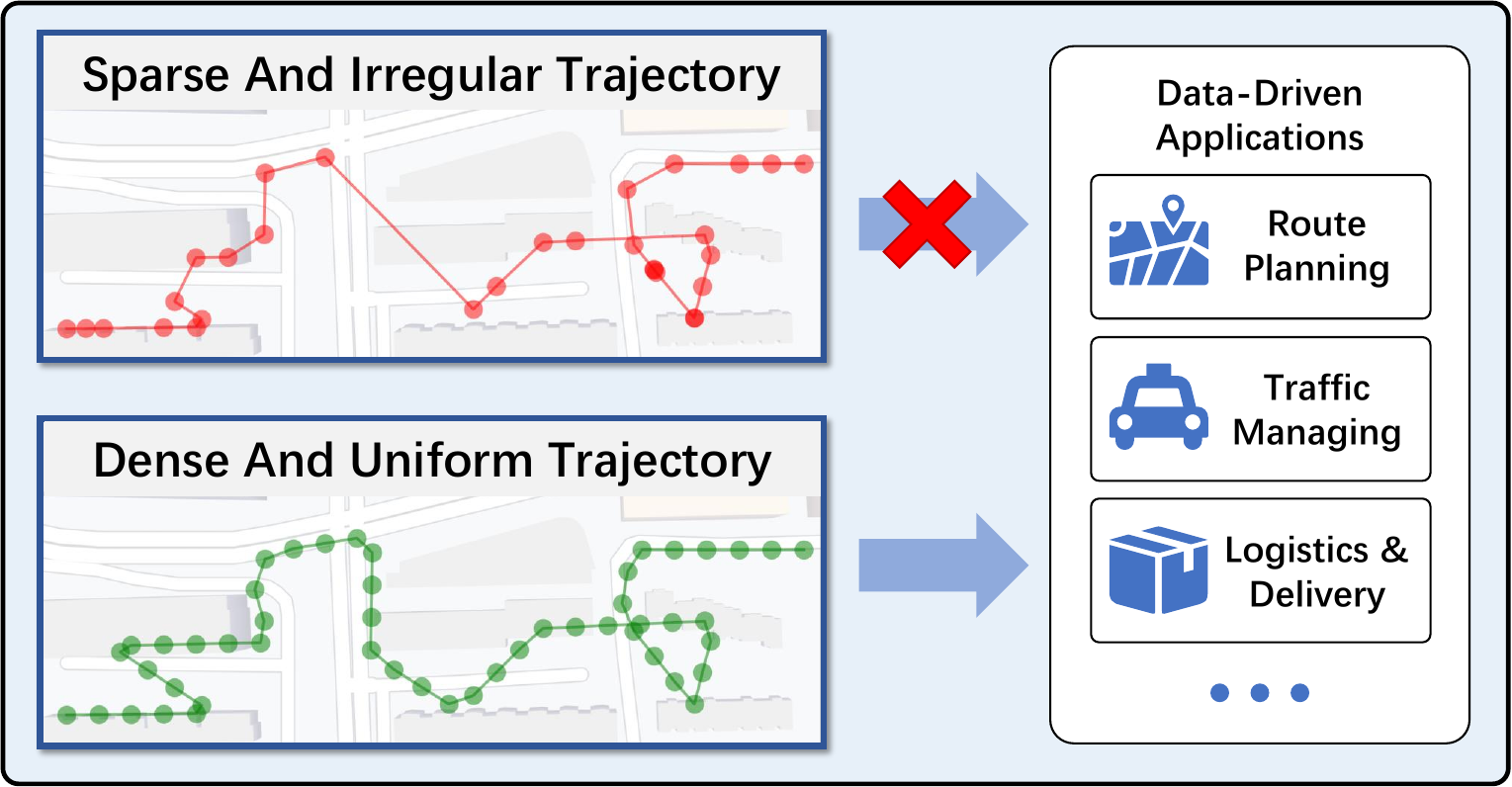}
    \caption{The trajectories collected from mobile edge devices are usually irregular and sparse, which fail to precisely describe the behavior of moving agents, thus cannot be used in data-driven applications.}
    \Description{Irregular and sparse trajectories cannot be used in downstream data-driven applications, such as route planning, traffic managing, logistics and delivery applications.}
    \label{fig:intro}
\end{figure}

%% file: 2_Preliminaries.tex
\section{Preliminaries}

\subsection{Problem Formulation}

\noindent\textbf{Timestamps:} Let $S = \{s_1, s_2, ... s_N\}$ denote a set of $N$ timestamps where $s_i \in \mathbb{R}^+$ represents the exact time of a GPS recording. Timestamps are ordered ($s_1 < s_2 < ... < s_N$) but are usually irregularly spaced, reflecting real-world constraints such as energy-saving sampling or signal loss.

\noindent\textbf{Trajectory:} Given timestamps $S$, a corresponding trajectory $\tau^S = \{p_{s_1}, p_{s_2}, ... p_{s_N}\}$ is a sequence of GPS coordinates, where each point $p_{s_i}$ records the longitude and latitude of the moving agent at time $s_i$ be a location point recording longitude and latitude.

\noindent\textbf{Sample Interval:} For consecutive timestamps $s_i$ and $s_{i+1}$, the sample interval $\Delta s_i= s_{i+1}-s_i$ quantifies temporal sparsity. Small and constant sample intervals computed from $S$ and $\tau^S$ indicate a dense and uniform trajectory, otherwise the trajectory is sparse and irregular.

\noindent\textbf{Query:} The query $Q=\{q_1, q_2, ... q_M\}$ specifies $M$ timestamps where the corresponding GPS coordinates $\{p_{q_1},p_{q_2}, ... p_{q_M}\}$ are to be inferred. Note that these coordinates form a complementary trajectory denoted as $\tau^Q$. Non-trivial trajectory recovery occurs when $Q\cap S = \emptyset$, requiring inference of locations at unobserved times.

\noindent\textbf{Trajectory Recovery:} Given a trajectory $\tau$ with large or irregular sample intervals, its corresponding $S$, a query $Q$, and other types of auxiliary contexts $\mathcal{C}$ such as date, user ID and waybills. The objective of the trajectory recovery task is to find a function that produces prediction $\hat{\tau}^Q$ and minimizes the error between $\hat{\tau}^Q$ and the ground-truth $\tau^Q$.

\input{__Figure_Main}

\subsection{Diffusion Models}

Denoising diffusion probabilistic models (DDPM)~\cite{DDPM} is constructed on the foundation of a forward Markov chain, which progressively corrupts clean data by adding Gaussian noise at each discrete step. The model trains a denoising neural network to accurately predict the noise introduced at each step of the forward process. Once trained, the network can be used to generate clean data such as trajectory points, starting from pure Gaussian noise.

Given a clean data sample $\mathbf{x}_0$ drawn from the training distribution $\mathbf{x}_0 \sim p_{\text{data}}(\mathbf{x})$. The forward Markov process is defined by a noise schedule $\beta_1, \beta_2, \dots, \beta_T$, specifying the variance of Gaussian noise added at each time step. At step $t$, the noisy data $\mathbf{x}_t$ is obtained through a conditional distribution: 
\begin{align}\label{eq:x_t_to_tp1}
\begin{split}
    q(\mathbf{x}_t | \mathbf{x}_{t-1}) &= \mathcal{N}(\mathbf{x}_t; \sqrt{1 - \beta_t} \mathbf{x}_{t-1}, \beta_t \mathbf{I}) \\
    x_{t}&=\sqrt{\alpha_t}x_{t-1}+\sqrt{\beta_t}\varepsilon_{t-1:t}
\end{split}
\end{align}

\noindent where $\alpha_t = 1 - \beta_t$ and $\varepsilon_{t:t+1} \sim \mathcal{N}(0,1)$ is single-step noise added. Alternatively, $\mathbf{x}_t$ can be expressed directly in terms of the clean data initial $\mathbf{x}_0$ as:

\begin{align}\label{eq:x_0_to_tp1}
\begin{split}
    q(\mathbf{x}_t | \mathbf{x}_0) &= \mathcal{N}(\mathbf{x}_t; \sqrt{\bar{\alpha}_t} \mathbf{x}_0, (1 - \bar{\alpha}_t) \mathbf{I}) \\
    x_{t}&=\sqrt{\Bar{\alpha}_t}x_0+\sqrt{1-\Bar{\alpha}_t}\varepsilon_{0:t}
\end{split}
\end{align}

\noindent where $\bar{\alpha}_t = \prod{i=1}^t\alpha_i$, the notation $\varepsilon_{0:t} \sim \mathcal{N}(0,1)$ is the multi-step noise, and $\mathbf{x}_1, \mathbf{x}_2, \dots, \mathbf{x}_T$ represent the intermediate states. As $T$ becomes sufficiently large, the forward process effectively transforms the data distribution $p_{\text{data}}(\mathbf{x}_0)$ into a standard Gaussian distribution.

To enable the reverse process, the model learns a denoising network parameterized by $\boldsymbol{\theta}$, which approximates the reverse transition distribution $p_{\boldsymbol{\theta}}(\mathbf{x}_{t-1}|\mathbf{x}_t)$ by minimizing the divergence from the true posterior $q(\mathbf{x}_{t-1}|\mathbf{x}_t, \mathbf{x}_0)$. This is achieved by training a noise predictor $\boldsymbol{\varepsilon}_{0:t, \theta}(\mathbf{x}_t, t)$, which estimates the noise $\boldsymbol{\varepsilon}_{0:t}$ introduced at each step.

During inference, the generation process involves iterative sampling of the learned reverse transitions $p_{\boldsymbol{\theta}}$, guided by the denoising network $\boldsymbol{\varepsilon}_\theta$. Additionally, the framework can incorporate conditional signals to improve generation quality. For instance, when conditioning on auxiliary information $\mathbf{c}$, the model can be extended to train a conditional denoising network $\varepsilon_{0:t, \theta}(\mathbf{x}_t, t, \mathbf{c})$ using an analogous loss function, thus enhancing the fidelity and control of the generated data.

There exist other formulations of diffusion processes that attempt to improve sampling efficiency through fewer generation steps. Such models include DDIM~\cite{DDIM}, score-based models~\cite{song2020score}, or PNDM~\cite{PNDM}. However, as long as the model still involves a multi-step denoising procedure, the redundancy persists, and they can be used jointly with our \diffusionname{}.

%% file: __Figure_Main.tex
\begin{figure*}[!ht]
  \begin{center}
  \includegraphics[width=\textwidth]{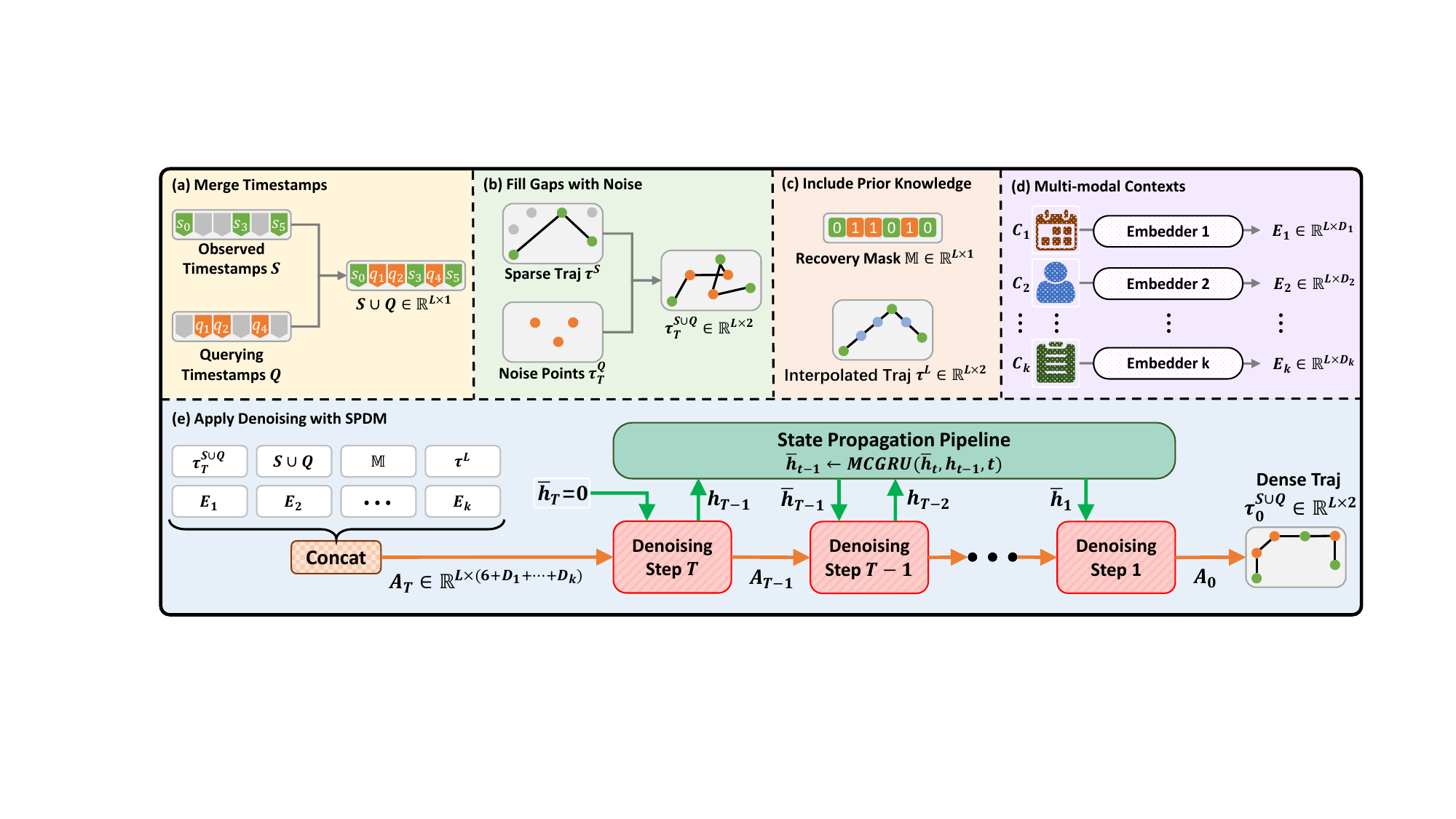}
  \caption{\methodname{} Framework. (a) Merge querying timestamps into observed timestamps in chronological order. (b) Initialize missing points with Gaussian noise. (c) Prior knowledge are included to explicitly tell the model where to recover. (d) The embedding of multi-modal contextual information. (e) The denoising process with SPDM that generates the dense trajectory.}
  \Description{TRACE first embeds all contextual data, then aggregate them into a single input tensor, finally underwent denoising process, which produces the dense and regular trajectory.}
  \label{fig:Main}
  \end{center}
\end{figure*}

%% file: 3_Methodology.tex
\section{Methodology}

\subsection{\methodname{} Overview}\label{sec:Method_Overview}

Trajectory recovery requires the integration of heterogeneous data, including trajectories, timestamps, contextual metadata (e.g., date and trip duration), and other user-specific features, which must be processed to a unified representation compatible with diffusion frameworks. These input data --- serving as the condition for trajectory recovery --- are illustrated in Fig.~\ref{fig:Main}(a-d), and the details will be discussed in Sec.~\ref{sec:Aggregation}. The aggregated representation is then fed into the denoising process, shown in Fig.~\ref{fig:Main}(e), which iteratively reconstructs the querying locations by removing noise from an initial Gaussian noise. Crucially, \methodname{} replaces the conventional denoising pipeline with \diffusionname{} to address the spatio-temporal irregularity problem. It introduces a state propagation mechanism that propagates multi-scale spatio-temporal features across denoising steps. The architecture of \diffusionname{}, including the specially designed neural networks used in each denoising step and the state propagation pipeline, will be explained in Sect.~\ref{sec:SPDM}. Lastly, the training procedure for \methodname{} is very different from the training of traditional diffusion models. This difference stems from the unique inter-step dependency introduced by the state propagation pipeline, and Sec.~\ref{sec:training} will present the details of new training algorithm.

\subsection{Condition Aggregation Stage}\label{sec:Aggregation}

To enable robust trajectory recovery under spatio-temporal irregularities, we integrate various prior knowledge to guide the recovery process. The sparse trajectory $\tau^S$ and associated timestamps $S$ are the primary observation that the recovery process can rely on, while the query timestamps $Q$ specifies target timestamps with unknown coordinates $\tau^Q$, which is initialized as Gaussian noise. The query timestamps can be manually or procedurally selected, ensuring highly flexible and customizable recovery pattern. To synchronize observation and query, we merge $S$ and $Q$ into a chronologically ordered sequence denoted $S\cup Q$ of length $L$. Similarly, we construct an assembled trajectory $\tau^{S\cup Q}$ following the same chronological order. To explicitly indicate which coordinates in $\tau^{S\cup Q}$ should undergo recovery, we label the missing point indices with a binary mask $\mathbb{M} \in \{0, 1\}^{L}$, where $\mathbb{M}_k=1$ indicates that the $k$-th timestamp in $S\cup Q$ comes from $Q$ and the $k$-th coordinate in $\tau^{S\cup Q}$ comes from $\tau^Q$. Moreover, we realize that simple linear interpolation is sufficient to recover some trajectory segments, e.g., straight paths, smooth and dense trajectory segments. Therefore, we introduce a linearly interpolated trajectory $lerp(\tau)$ over $S\cup Q$ to augment the input. Concretely, it fills the unknown coordinates with linearly interpolated values inferred from nearby points, providing the model with a geometrically plausible prior.

Industrial applications usually provide a variety of heterogeneous contexts $\{C_1, C_2, ... C_K\}$ associated with trajectories, e.g. weekday, date, user ID, waybill details. Since the embedding part is not the focus of our method, we assume that one module is specially designed for each context, where each module consists of several simple operations such as linear or convolutional layers. In the end, each context $C_i$ is converted to a $d_i$ dimensional sequential feature $E_i\in\mathbb{R}^{L\times d_i}$ aligned with the trajectories. 

At the end of this stage, the final aggregated feature tensor $A\in\mathbb{R}^{L\times D}$ is constructed as:
\begin{equation}\label{eq:Aggragate}
    A \gets Concatenate(\tau^{S\cup Q}, S\cup Q, \mathbb{M}, lerp(\tau), E_1, ... E_K)
\end{equation}
\noindent where $D$ is the final dimensionality of the features. To conclude, this formulation ensures sparse and irregular query regions, enriched with various prior knowledge and contextual information, enabling adaptive recovery of complex trajectory segments.

\subsection{Denoising With SPDM}\label{sec:SPDM}

Traditional diffusion models iteratively denoise the input noise $x^T$ over $T$ steps, transforming it into clean data $x^0$. In the proposed framework, our input is denoted as $A_T$, which will be recovered to $A_0$ through intermediate steps $\{A_{T-1}, A_{T-2} ... A_1\}$. In fact, although $A_t$ is the input for step $t$, the denoising operation only applies to $\tau_t^Q$, which is a portion inside $A_t$ corresponding to the noisy coordinates, while the other portion of $A_t$ is fixed throughout the entire denoising process.

A novel formulation for the denoising step is essential to enable effective state propagation in SPDM. We begin by defining the states: let $h_t$ be the \textbf{single-step} hidden state that encapsulates the local features extracted at step $t$, and let $\bar{h}_t$ be the \textbf{multi-step} hidden state containing all useful knowledge from previous steps $[t+1, T]$. To enhance feature richness and reusability, we define both single- and multi-step hidden states as multi-scale sequential features, i.e. $h_t = \{f_{t;1}, f_{t;2}, ... f_{t;b}\}$ and $\bar{h}_t = \{\bar{f}_{t;1}, \bar{f}_{t;2}, ... \bar{f}_{t;b}\}$, where $b$ is the number of sequential features in the state. Under this formulation, for each denoising step $t$, the denoising network takes $A_t$, $\bar{h}_t$ and $t$ as inputs and produces $h_{t-1}$ along with the predicted noise $\hat{\varepsilon}_{0:t}$. Subsequently, the state propagation network accepts $\bar{h}_t$, $h_{t-1}$ and $t$ to output the updated state $\bar{h}_{t-1}$.

Based on this formulation, our neural network architecture is implemented accordingly. The denoising network in \methodname{} is built upon a UNet architecture, which inherently supports multi-scale feature extraction and aligns well with our state definitions. To accommodate state propagation, the UNet is modified as illustrated in the upper part of Fig.~\ref{fig:model}. Specifically, the UNet comprises exactly $b$ basic blocks --- matching the number of features in the states. Building on top of traditional UNet blocks, each modified block $i$ also processes additional input $\bar{f}_{t;i}\in \bar{h}_t$ to generate the corresponding output feature $f_{t-1;i}$. Finally, all output features of the $b$ UNet blocks are assembled to $h_{t-1}$.

After we get $h_{t-1}$ generated by the denoising UNet, the state propagation network is responsible for updating the cumulative knowledge and producing $\bar{h}_{t-1}$. Concretely, we implement a novel Multi-scale Convolutional Gated Recurrent Unit (MCGRU) with diffusion step awareness, shown in the lower part of Fig.~\ref{fig:model}. MCGRU contains $b$ CGRU blocks in parallel, mirroring the UNet's hierarchy. Building on top of traditional CGRU, the modified version also takes into account the diffusion time $t$, which enhances its compatibility with the multi-step denoising pipeline.

\input{__Figure_Model}

\subsection{Training of \methodname{}}\label{sec:training}

In traditional diffusion model training for trajectory recovery, a dense trajectory with corresponding timestamps is provided, which are then divided into the observation part ($S$, $\tau^S$) and the query part ($Q$, $\tau_0^Q$). With a randomly selected time step $t\in[1, T]$, the forward diffusion process is applied to generate $\tau_t^Q$ by adding noise $\varepsilon_{0:t} \in \mathcal{N}(0, 1)$. Then, the denoising network is trained to predict $\varepsilon_{0:t}$. However, during \methodname{} training, the \diffusionname{} denoising step $t$ requires the state $\bar{h}_{t}$ from previous iterations. This disables random step sampling and necessitates a sequential training paradigm, where each data sample iterates through all steps from $t=T$ to $t=1$, so that $\bar{h}_{t}$ is always available.

The above solution makes the training of the denoising network possible, yet we also have to train the state propagation network, i.e., each state $\bar{h}_t$ must contain useful knowledge for the subsequent steps. This requirement imposes two sub-constraints: i) the denoising network at step $t$ should output appropriate $h_{t-1}$ and ii) the state propagation network should effectively aggregate or filter knowledge to yield beneficial $\bar{h}_{t-1}$. Intuitively, the ideal strategy is to jointly train the entire denoising process, with the overall objective being:

\begin{equation}
  \min_{\theta}\sum_{t=1}^T MSE(\bar{\varepsilon}_{0:t}, \varepsilon_{0:t})
\end{equation}

\noindent where $\theta$ represents the learnable parameters of the models. However, incorporating all tens to hundreds of denoising steps in each training iteration is computationally impractical. Instead, we partition the denoising process into smaller segments comprising several consecutive steps, e.g. including two steps in one training iteration:

\begin{equation}\label{eq:arch_obj_2step}
  \min_{\theta} (MSE(\hat{\varepsilon}_{0:t}, \varepsilon_{0:t}) + MSE(\hat{\varepsilon}_{0:t-1}, \varepsilon_{0:t-1}))
\end{equation}

Another challenge in training \methodname{} arises from the noise sampling process. The standard diffusion model training algorithm samples $\varepsilon_{0:t} \sim \mathcal{N}(0, 1)$ independently in each training iteration, because it does not involve the complex inter-step dependencies introduced by the state-propagation mechanism. In contrast, our algorithm iterates $t$ from $T$ to $1$ for each data sample, and one training iteration involves at least two denoising steps. Consequently, the training of \methodname{} cannot ignore the dependency among noises, i.e., multi-step noises $\varepsilon_{0:t}$ and $\varepsilon_{0:t-1}$. To derive a set of dependent noise samples, we first sample a list of single-step noises $\{\varepsilon_{0:1}, \varepsilon_{1:2}, ..., \varepsilon_{T-1:T}\}$, where each $\varepsilon_{t-1:t} \sim \mathcal{N}(0, 1)$ is always independent. Then, starting with $A_0$, we recursively apply single-step diffusion forward using Eq.~\ref{eq:x_t_to_tp1} to obtain $A_1$ to $A_T$. Next, we apply following equation to compute the multi-step noise $\varepsilon_{0:t}$ for each $t$:

\begin{equation}\label{eq:get_noise}
  \varepsilon_{0:t} = \frac{A_t - \sqrt{\bar{\alpha}_t} A_0}{\sqrt{1 - \bar{\alpha}_t}}
\end{equation}

\noindent This equation is a changed form of Eq.~\ref{eq:x_0_to_tp1}, it ensures that the dependency among the noises is respected within each training iteration, as well as during the entire life cycle of a data sample. Now, the general principles of training \methodname{} have finally been established, and a naive procedure of the proposed training algorithm is presented in Algorithm~\ref{algo:training}.

\input{__Algo_Training}

Although the above design is sufficient for training, practical implementation demands additional techniques. As mentioned above, each data sample in the training batch undergoes $T$ training iterations with a synchronized countdown of $t$ across the entire batch (Fig.~\ref{fig:batch_loading}, top). Consequently, the model overfits to a narrow range of denoising steps and cannot generalize to the full range of $t$. To address this, we implement a dynamic batch management technique that assigns a private $t$ to each sample in a batch, with an offset among them (Fig.~\ref{fig:batch_loading}, middle). We can further promote generalization across the entire range of $t$ by uniformly assigning $t \in [1, T]$ to each data sample, as illustrated in Fig.~\ref{fig:batch_loading} bottom. In this scheme, the $t$ of each sample is updated independently, and a new sample is loaded in-place once its $t$ reaches 0.

\input{__Figure_Batch}

%% file: __Figure_Model.tex
\begin{figure}
    \centering
    \includegraphics[width=1.0\linewidth]{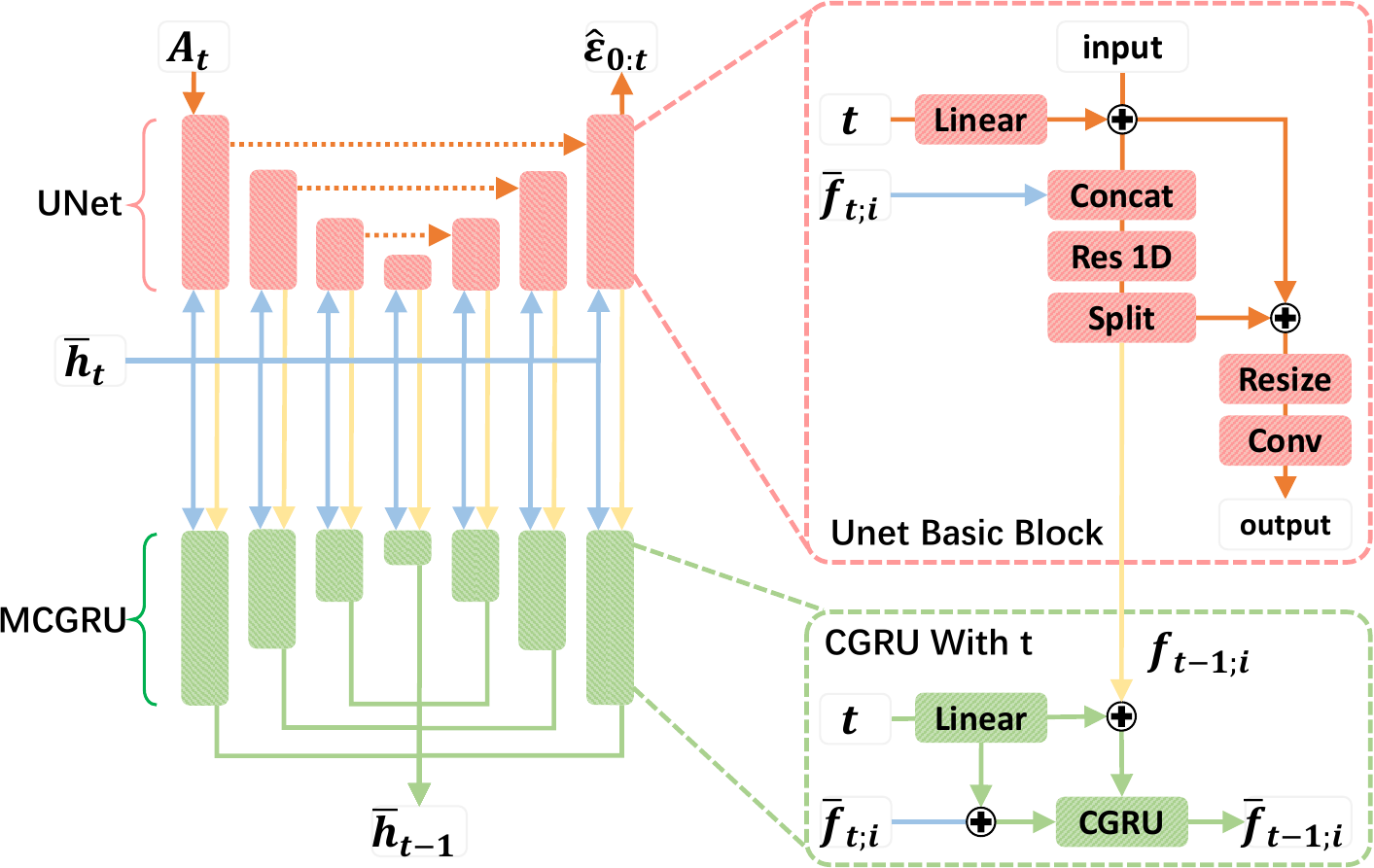}
    \caption{The architecture of the modified UNet (upper, red) and the proposed MCGRU (lower, green).}
    \Description{In the proposed UNet architecture, each UNet stage takes state features as an additional input, then produces the processed state features, which serve as the input for the next denoising step.}
    \label{fig:model}
    \vspace{-5mm}
\end{figure}

%% file: __Algo_Training.tex
\begin{algorithm}[t]
\caption{Training One Sample (2-step)}
\label{algo:training}

\begin{algorithmic}[1]
    \item \textbf{Input}: $S,\ Q,\ \tau^S,\ \tau_0^Q,\ lerp(\tau),\ \mathbb{M},\ C_1,\ C_2,\ \cdots\ C_K$.
    \STATE Generate $\{\varepsilon_{t-1:t} | t \in [1, T]\}$.
    \STATE Derive $\{\varepsilon_{0:t} | t \in [1, T]\}$ with Eq.~\ref{eq:get_noise}.
    \STATE Apply diffusion to $\tau_0^Q$ to get $\{\tau_t^Q | t \in [1, T]\}$.
    \STATE $h_T \gets 0$
    \FOR{$t \gets T$ to $1$}

        \STATE $E_i \gets Embed_i(C_i)$ for $i\in\{1,...K\}$.

        \STATE Obtain $A_{t-1}, A_{t}$ using concatenation as in Eq.~\ref{eq:Aggragate}.
        
        \STATE \textbf{// Train the first step}
        \STATE $\hat{\varepsilon}_{0:t}, h_{t-1} \gets$ UNet$(A_t, \bar{h}_t, t)$

        \STATE \textbf{// Update hidden state}
        \STATE $\bar{h}_{t-1} \gets$ MCGRU$(\bar{h}_t, h_{t-1}, t)$

        \STATE \textbf{// Train the second step}
        \STATE $\hat{\varepsilon}_{0:t-1}, h_{t-2} \gets$ UNet$(A_{t-1}, \bar{h}_{t-1}, t-1)$
        \STATE Compute loss with Eq.~\ref{eq:arch_obj_2step}. and do optimization.
    \ENDFOR
\end{algorithmic}
\end{algorithm}

%% file: __Figure_Batch.tex
\begin{figure}
    \centering
    \includegraphics[width=0.9\linewidth]{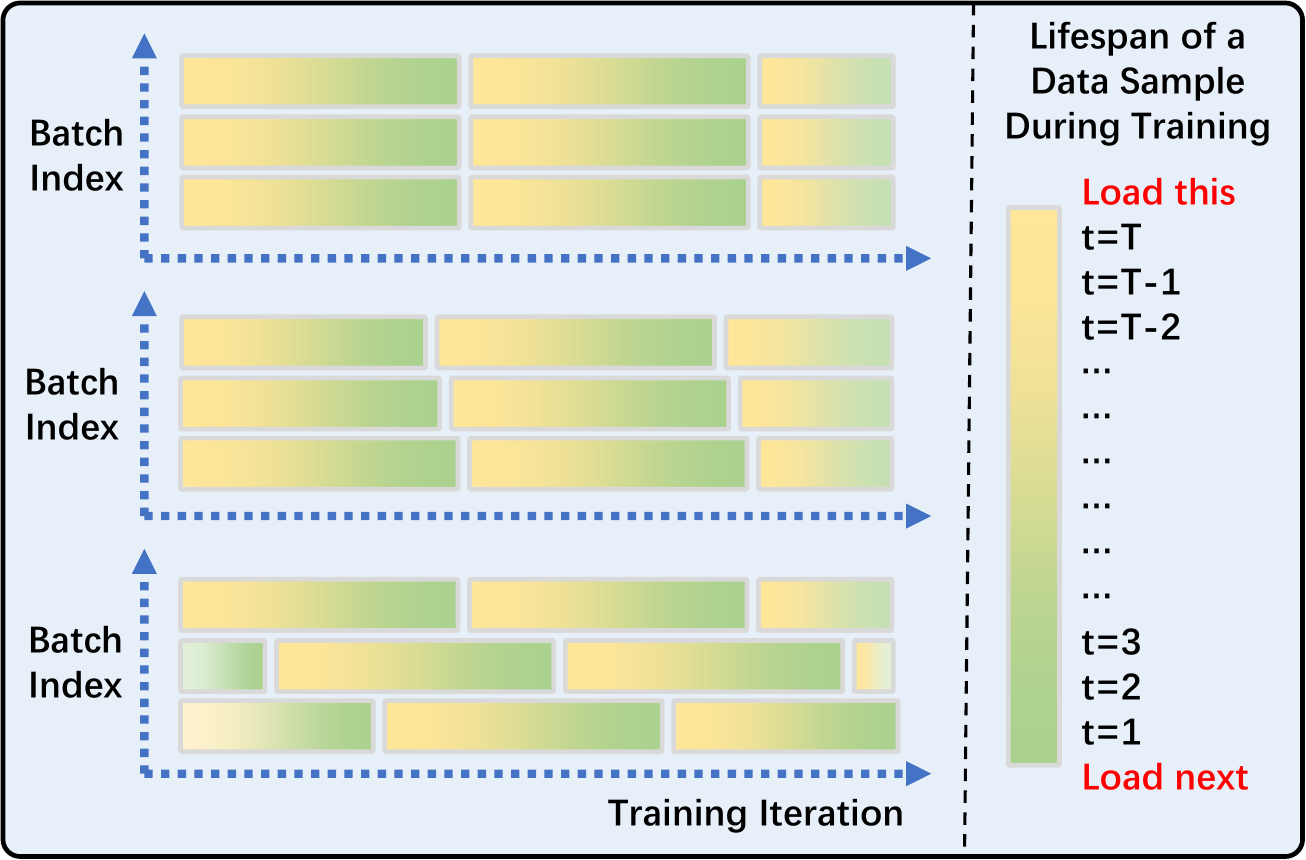}
    \caption{Three ways of batch management. Top: Shared $t$ among all samples in a batch. Middle: Different $t$ for each sample with little offset. Bottom: Uniformly distributed $t$ over the range $[1, T]$.}
    \Description{A single data sample (trajectory) will be used many times consecutively, if we load the entire batch of data with the same $t$, the model training become unstable. In contrast, try to make $t$ evenly distribution among data samples in a batch is better.}
    \label{fig:batch_loading}
    \vspace{-5mm}
\end{figure}

%% file: 4_Experiment.tex
\section{Experiments}

\subsection{Experimental Settings}

\paragraph{Datasets.}
We evaluate \methodname{} on three datasets. Two of them comprise dense and smooth taxi trajectories collected in Xi'an city and Chengdu city, China. To assess the model's performance across diverse mobility patterns and metadata, we also include a dataset from last-mile logistics, capturing courier movements in residential areas with irregular sampling intervals. For conditional contexts and metadata, we extract the trajectory starting date and time, the moving agent's ID, and trip duration. In addition, the logistics data set includes information for package delivery and receiving. All GPS coordinates undergo z-score normalization, and the time stamp sequence of each trajectory is rescaled to the range $[0, 1]$.

\input{__Table_Overall}

\input{__Figure_Irregularity}

\paragraph{Metrics.}
Trajectory recovery focuses on the exact spatial similarity between the recovered coordinates and the ground truth ones. Therefore, we adopt three evaluation metrics: The Mean Squared Error (MSE), the Mean Absolute Error (MAE), and the Normalized Dynamic Time Warping (NDTW). Lower values are preferable for all metrics.

\paragraph{Baselines.}
We evaluate \methodname{} against four baselines. i) \textbf{DeepMove}~\cite{DeepMove}: An RNN-based trajectory forecasting model adapted for recovery via auto-regressive imputation. ii) \textbf{AttnMove}~\cite{AttnMove}: A Transformer-based approach fusing intra-/inter-trajectory contexts via self- and cross-attention. iii) \textbf{PriSTI}~\cite{PriSTI}: A diffusion model for time-series imputation, modified for trajectories by treating GPS coordinates as 2D signals. iv) \textbf{DT+RP}: Combines DiffTraj~\cite{DiffTraj} (UNet-based trajectory generation) with RePaint~\cite{ImageInpainting} (image inpainting via diffusion), which is capable of adapting content generation diffusion models to recovery tasks. Additionally, we ablate \methodname{} by replacing \diffusionname{} with a standard diffusion pipeline (\textbf{TRACE w/o SPDM}), which aims to test the effectiveness of \diffusionname{}.

\subsection{Results}

\subsubsection{Overall Comparison.}

Table~\ref{tab:overall} summarizes the performance of \methodname{} against the baselines. Non-diffusion methods (DeepMove, AttnMove) exhibit significantly higher errors, underscoring the limitations of auto-regressive and attention-based architectures in handling sparse, irregular trajectories. In contrast, Diffusion-based baselines show marked improvements except for the approach combining DiffTraj with RePaint, which emphasizes the need to design and train diffusion models specific for recovery tasks. The ablation study (\methodname{} without \diffusionname{}) highlights the need for state propagation, while competitive, it underperforms \methodname{} in most test cases, highlighting the importance of \diffusionname{}.

\subsubsection{Addressing Spatio-Temporal Irregularity.}
We prove the effectiveness of \diffusionname{} on addressing the challenge of spatio-temporal irregularity inherent to trajectory data. Given a trajectory to be recovered $\tau^S$, we compute two metrics to measure the spatial and temporal irregularities: i) the standard deviation of distances between consecutive point pairs and ii) the standard deviation of sample intervals. A large variance indicates that the trajectory is irregular and skewed; otherwise, the trajectory points are uniformly distributed. The results are shown in Fig.~\ref{fig:Irregularity}. As the trajectory gets increasing  irregular, all methods tend to have worse performance. However, the proposed \methodname{} is apparently more robust to such irregularities. When comparing \methodname{} with and without \diffusionname{}, we can see that the \diffusionname{} achieves significant improvement when dealing with irregular trajectories, showcasing the effective and successful design of \diffusionname{}.

\input{__Figure_Scalability}

\input{__Table_Efficiency}

\subsubsection{Scalability.}
The length and sparsity of trajectories can vary greatly in industrial scenarios. In this experiment, we study the scalability of the proposed method with different trajectory lengths and sparsity levels. As illustrated in Fig.~\ref{fig:Scalability}, we first select the three strongest methods: PriSTI, \methodname{} without \diffusionname{}, and \methodname{}. The left column presents three error metrics with trajectory length varying from 64 to 512, and the right column shows the slowly increasing errors as the trajectory gets more sparse. The sparsity is measured by the percentage of points removed from a dense trajectory. As we can see, the plots exhibit parallel trends, while \methodname{} consistently outperforms other methods. 

\subsubsection{Efficiency.}
DDIM~\cite{DDIM} is a variant in the diffusion model family that can speed up recovery by skipping some intermediate denoising steps, trading result quality for faster recovery. We compare the recovery time and quality of \methodname{} using DDIM with or without state propagation to demonstrate the superior efficiency of \methodname{}, and the results are shown in Table \ref{tab:compare_time_quality}. The denoising times are recorded with 100 samples in a batch on a NVIDIA GeForce RTX 3070 Ti Laptop GPU. For plain DDIM, the 11 denoising steps version is $119.176/3.18=37.48$ faster than the full 500-step version, but the recovery quality is decreased by $(0.0490-0.0432)/0.0490\times100\%=-11.84\%$. In contrast, building a state propagation pipeline on top of the 11-step DDIM introduces only 12\% more parameters and 0.4 seconds of running time, and also improves the upper bound of recovery quality by $(0.0349-0.0256)/0.0349\times100\%=26.65\%$. This indicates that the proposed \diffusionname{} can be jointly used with other diffusion sampling techniques that accelerate recovery, while achieving even higher recovery quality.

We also provide a table describing the computational cost of all baselines, shown as Table~\ref{tab:flops}. In particular, we present the number of parameters,  multiply-accumulate operations (MACs), and floating point operations (FLOPs) of all baselines. We notice that \methodname{} is very parameter-efficient due to its efficient design. The SPDM only introduces a small number of parameters and computations to our diffusion model, this further proves its effectiveness. One thing to note is that the diffusion models require iteratively denoising, so the cost should be multiplied by the denoising steps, making them computationally costly compared to non-diffusion-based baselines.

\subsubsection{Effectiveness of Training Algorithms.}
We analyze the effectiveness of the proposed training algorithms and techniques introduced in Sect.~\ref{sec:training}. First, we want to study the effect of different number of denoising steps included in a single training iteration, as shown in Fig.~\ref{fig:comp_step}. The result indicates that the number of steps trained in each iteration do not have a great influence on the final performance, and 2-step is preferable since it requires the least training time. Second, we show the importance of dynamic batch management during training. As we can see in Fig.~\ref{fig:comp_batching}, the uniformly distributed $t$ for each data sample within a batch leads to the best convergence, while a shared $t$ for the entire batch results in the worst recovery quality. This study highlights the necessity of the proposed training algorithm.

\input{__Figure_Ablation}

\subsection{Case Study}

\input{__Table_CaseStudy}

\input{__Figure_Visualization}

A case study was conducted using courier trajectories in a logistics scenario. Analyzing the speed and distance of moving people or vehicles is crucial for many location-based data-driven Web applications, such as navigation and last-mile delivery. The trajectory of the user is usually collected to compute the speed and the moving distance. However, these trajectories are usually sparse due to constrained device-server communication rates, limited sampling rates, and unstable signal conditions. In addition, storing massive amounts of trajectories for large-scale web-service can be costly. All these limitations hinder accurate computation of user speed and moving distance.

We collected very dense trajectories in a crowded apartment region. A subset was sampled from each dense trajectory according to a normal last-mile delivery sampling interval, resulting in only 30\% of the points remaining. \methodname{} was trained to recover these sparse trajectories to dense ones. We used the moving distances and speed computed from the very dense trajectories as ground truth and then computed the estimation using the sparse and the recovered trajectories. For more direct comparison, we normalized the ground truth speed to 1 meter per second and the moving distance to 1 kilometer, and the results are shown in Table \ref{tab:case_study_score}. We observed that the estimations using sparse trajectories captured 60\% of the actual speed and distance, while the recovered trajectories increased this percentage to around 83\%. An illustration of the trajectory before and after recovery is shown in Figure \ref{fig:case_study_visual}.

%% file: __Table_Overall.tex
\begin{table}[t]
\centering
{\footnotesize
\begin{tabular}[t]{l|l|ccc}
\toprule
\multirow{3}{*}{Metric} & \multirow{3}{*}{Method} & \multicolumn{3}{c}{Dataset} \\
\cmidrule(r){3-5}
& & Xi'an & Chengdu & Logistics \\
\midrule
MSE $\downarrow$
  & DeepMove & 5.297 & 26.857 & 30.060\\
  ($\times 10^{-3}$) & {AttnMove} & 0.068 & 0.234 & 3.201\\
  & {PriSTI} & 0.019 & 0.292 & 2.206\\
  & {DT + RP} & 0.326 & 6.919 & 4.250\\
  & \textbf{\methodname{} w/o SPDM} & \snd{0.017}& \snd{0.202}& \snd{2.025}\\
  & \textbf{\methodname{}} & \fst{0.010}& \fst{0.159}& \fst{0.449}\\
  \midrule
NDTW $\downarrow$
  & {DeepMove} & 36.295 & 169.470 & 72.814\\
  ($\times 10^{-3}$) & {AttnMove} & 2.257 & 5.458 & 22.652\\
  & {PriSTI} & 1.174 & 3.975 & 12.656\\
  & {DT + RP} & 8.473 & 28.773& 13.006\\
  & \textbf{\methodname{} w/o SPDM} & \snd{1.156}& \snd{3.809}& \snd{11.062}\\
  & \textbf{\methodname{}} & \fst{1.154}& \fst{3.597}& \fst{5.520}\\
  \midrule
MAE $\downarrow$
  & {DeepMove} & 31.695 & 149.169 & 94.151 \\
  ($\times 10^{-3}$) & {AttnMove} & 3.119 & 7.250 & 33.744 \\
  & {PriSTI} & 1.587 & \snd{2.397} & \snd{2.233} \\
  & {DT + RP} & 3.195 & 112.212 & 21.458 \\
  & \textbf{\methodname{} w/o SPDM} &  \snd{1.352} & 4.646 & 21.796 \\
  & \textbf{\methodname{}} & \fst{1.335}& \fst{1.725} & \fst{2.121} \\

\bottomrule
\end{tabular}
}
\caption{The overall comparisons of six models with three metrics on three dataset. The trajectory length is 512 points, with 50\% GPS points erased and the models are asked to recover the erased points.}
\label{tab:overall}
\vspace{-3mm}
\end{table}

%% file: __Figure_Irregularity.tex
\begin{figure}[t]
\centerline{\includegraphics[width=\linewidth]{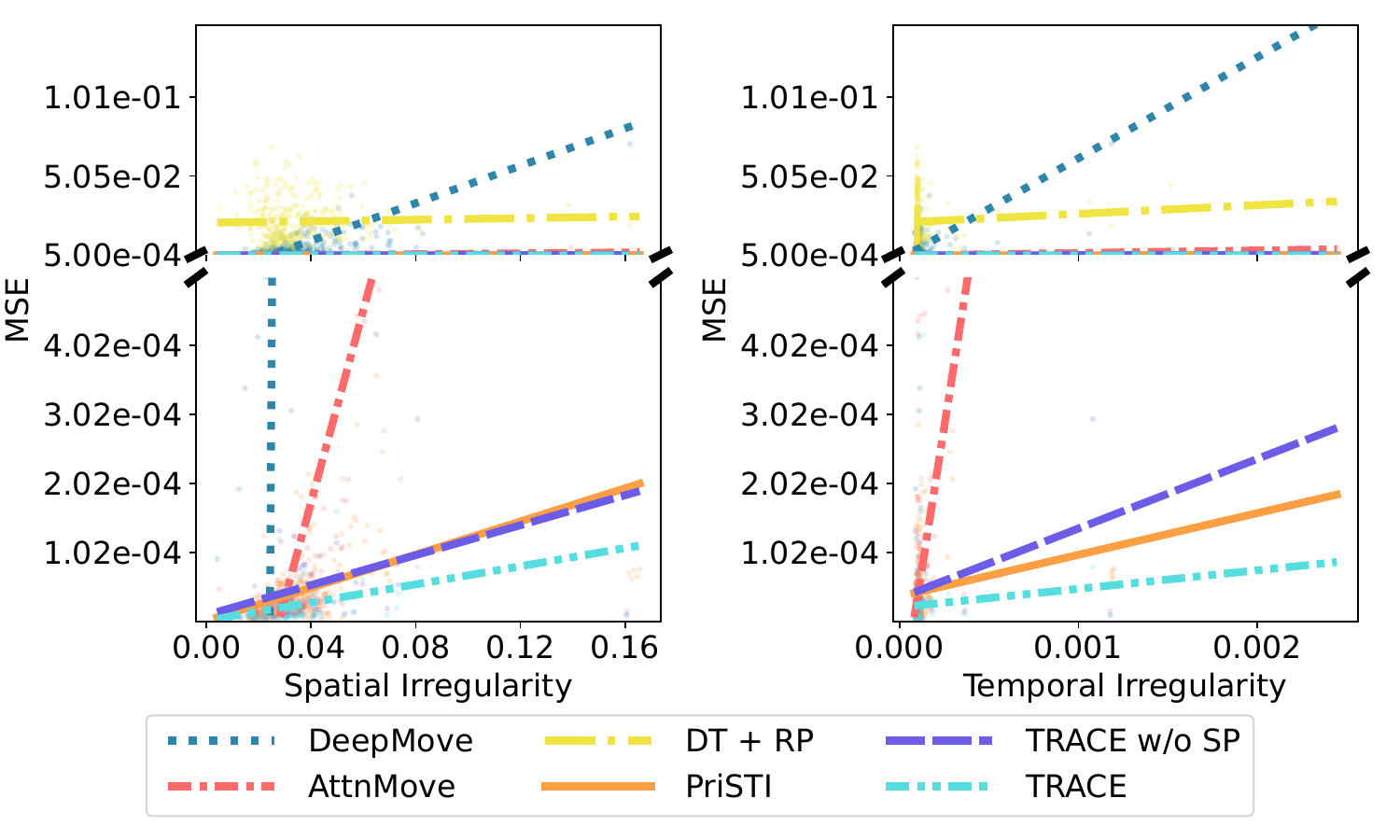}}
\caption{The trend of recovery errors as the trajectory gets more irregular in terms of spatial and temporal distributions. There are breaks in y-axis because the magnitude of MSE differs a lot for different methods. Tested on Xi'an with 70\% points erased and trajectory length of 512 points.}
\Description{The proposed model shows the best scalability as the trajectory gets more irregular in terms of spatial and temporal distributions.}
\label{fig:Irregularity}
\end{figure}

%% file: __Figure_Scalability.tex
\begin{figure}[t]
\centerline{\includegraphics[width=0.98\linewidth]{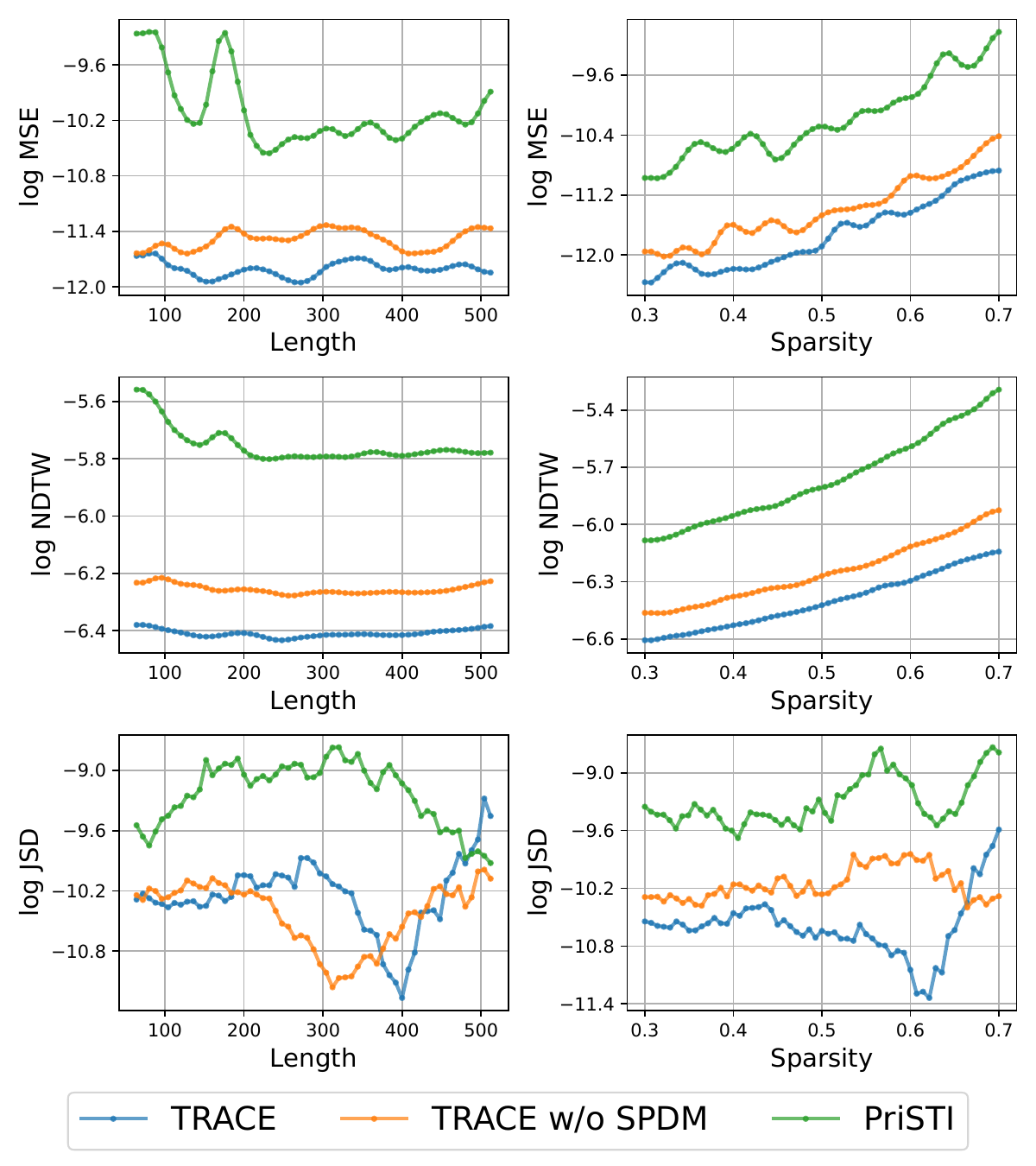}}
\caption{The comparison among sparsity levels and trajectory lengths.}
\Description{The proposed model (with SPDM) shows the best robustness across different trajectory lengths and sparsity levels.}
\label{fig:Scalability}
\end{figure}

%% file: __Table_Efficiency.tex
\begin{table}[t]
    \centering
    \setlength{\tabcolsep}{2.2mm}{
    \begin{tabular}{c|ccccc}
        \toprule
         \TwoRow{Method} & \TwoRow{Steps} & Time & MSE & NDTW & MAE \\
         & & (s) & $\times 10^{-3}$ & $\times 10^{-3}$ & $\times 10^{-3}$ \\
         \midrule
         & 500 & 119.176 & 0.0432 & 2.01 & 3.068 \\
         DDIM & 51 & 12.36 & 0.0349 & 2.868 & 2.221 \\
         & 26 & 6.65 & 0.0351 & 2.897 & 2.239 \\
         & 11 & 3.18 & 0.0490 & 3.083 & 2.365 \\
         \midrule
         DDIM & 500 & 129.57 & 0.0356 & 2.750 & 2.502 \\
         With & 51 & 13.02 & 0.0272 & 2.332 & 1.847 \\
         State & 26 & 7.03 & 0.0256 & 2.320 & 1.822 \\
         Prop & 11 & 3.58 & 0.0260 & 2.420 & 1.914 \\
         \bottomrule
    \end{tabular}
    }
    \caption{The comparison of recovery time and quality with and without state propagation pipeline. Tested on Xi'an with 70\% points erased and trajectory length of 512 points.}
    \label{tab:compare_time_quality}
    \vspace{-5mm}
\end{table}

\begin{table}[t]
    \centering
    \begin{tabular}{c|c|c|c}
        \toprule
        Method & Parameters & MACs & FLOPs \\
        \midrule
        \textbf{\methodname{}} & \snd{6.28M} & 1.13G & 2.28G \\
        \textbf{\methodname{} w/o SPDM} & \fst{6.17M} & \snd{0.99G} & \snd{2.01G} \\
        DT + RP & 9.10M & \fst{0.63G} & \fst{1.27G} \\
        PriSTI & 7.32M & 3.24G & 6.51G \\
        AttnMove & 8.15M & 3.94G & 7.92G \\
        DeepMove & 10.11M & 1.94G & 9.93G \\
        \bottomrule
    \end{tabular}
    \caption{The number of parameters, multiply-accumulate operations (MACs), and floating point operations (FLOPs) of all methods.}
    \label{tab:flops}
    \vspace{-3mm}
\end{table}

%% file: __Figure_Ablation.tex
\begin{figure}[t]
\centering
        \begin{subfigure}[t]{0.9\linewidth}
		\centering
		\includegraphics[width=\linewidth]{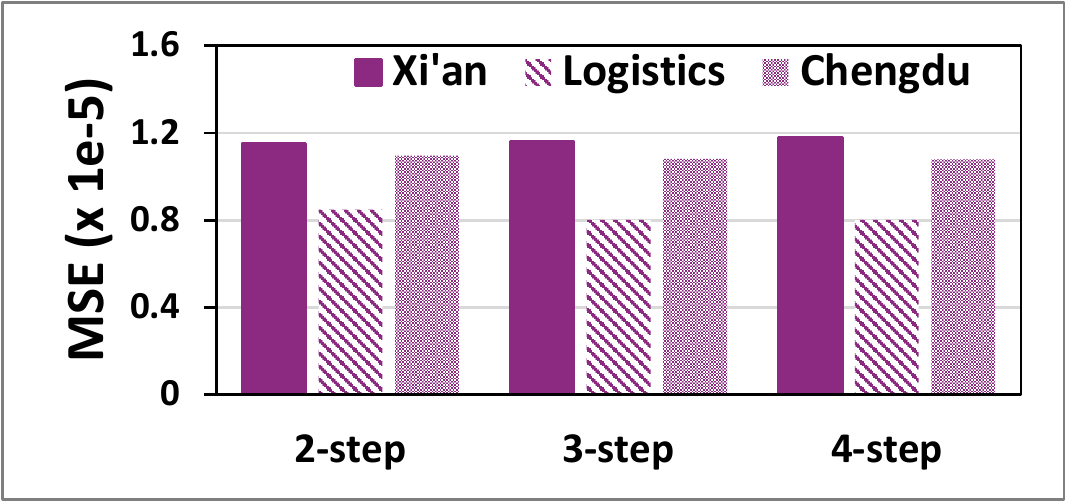}
        \subcaption{Steps in each iteration}
		\label{fig:comp_step}
	\end{subfigure}
	\begin{subfigure}[t]{0.9\linewidth}
		\centering
		\includegraphics[width=\linewidth]{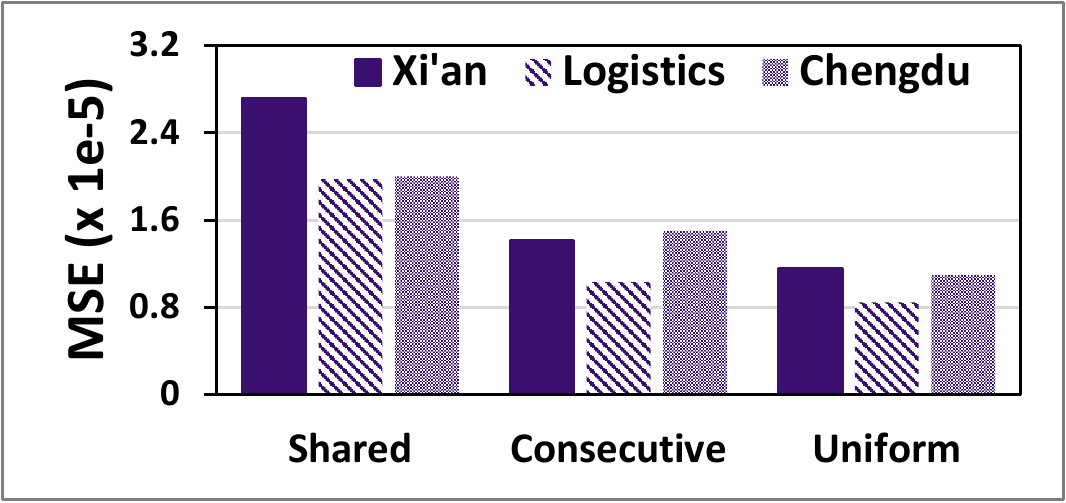}
        \subcaption{Batch managing methods}
		\label{fig:comp_batching}
	\end{subfigure}
 \caption{The comparisons among: (a) Number of denoising steps included in single training iteration. (b) Three ways of batch managing introduced in Fig~\ref{fig:batch_loading}. Tested on Xi'an with 70\% points erased and trajectory length of 512 points.}
 \Description{The number of denoising steps included in a single training iteration has very little effect on final performance. While the proposed batch management algorithm is necessary for a stable training and significantly reduces loss.}
 \label{fig:comp_linkage_algo}
\end{figure}

%% file: __Table_CaseStudy.tex
\begin{table}[t]
\centering
{
\begin{tabular}{l|c|c}
    \toprule
    & Average Speed & Moving Distance \\
    & (m/s) & (km) \\
    \midrule
    Ground Truth & 1 & 1 \\

    Sparse Traj & 0.6180 & 0.5788 \\

    Recovered Traj & 0.8252 & 0.8480 \\
    \bottomrule
\end{tabular}
}
\caption{The moving speed and distance estimations computed from trajectories after normalization.}
\label{tab:case_study_score}
\vspace{-8mm}
\end{table}

%% file: __Figure_Visualization.tex
\begin{figure}[t]
\centering
\begin{subfigure}[t]{0.49\linewidth}
    \centering
    \includegraphics[width=\linewidth]{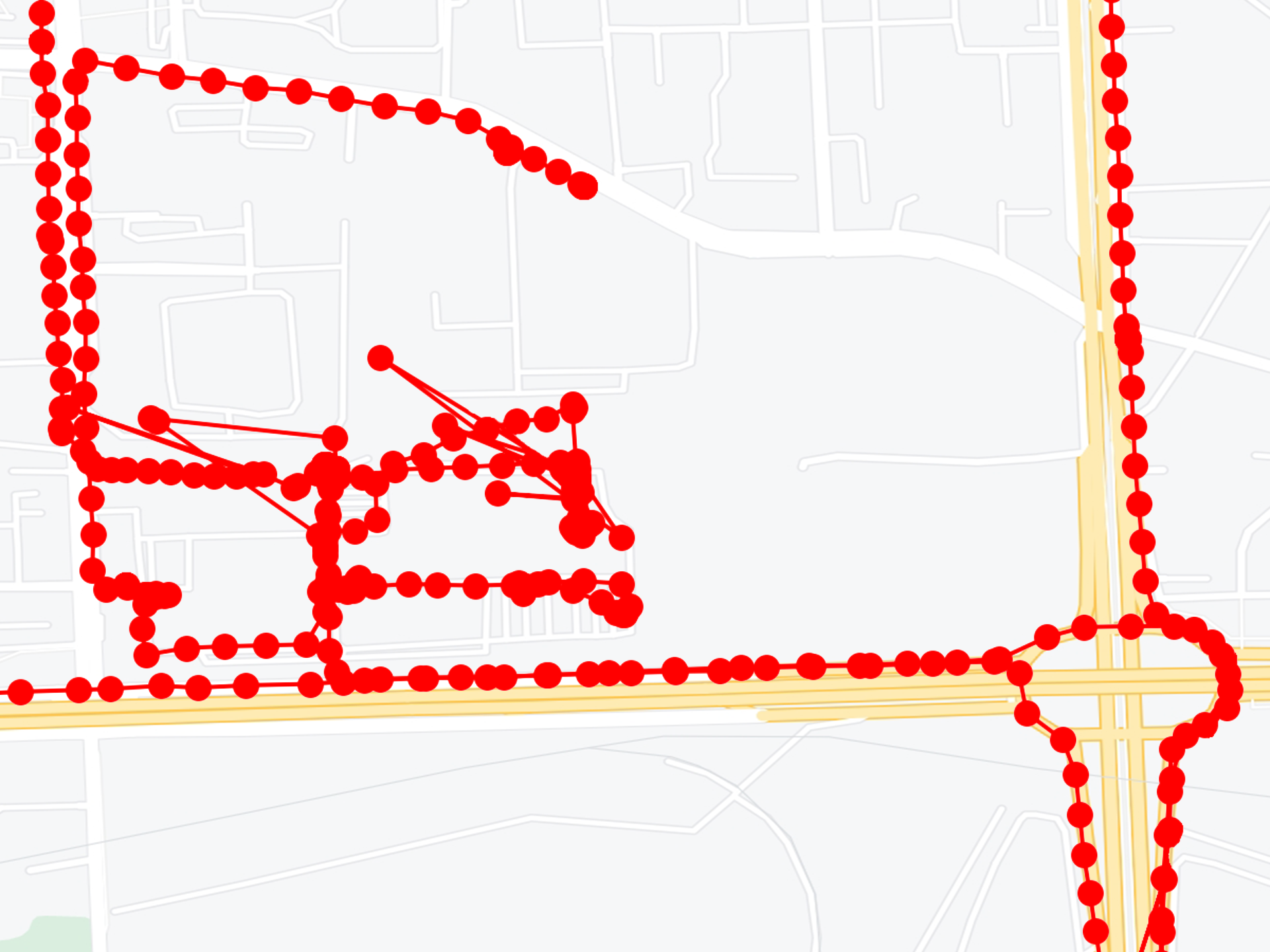}
    \subcaption{Ground truth trajectory}
    \label{fig:case_dense}
\end{subfigure}
\begin{subfigure}[t]{0.49\linewidth}
    \centering
    \includegraphics[width=\linewidth]{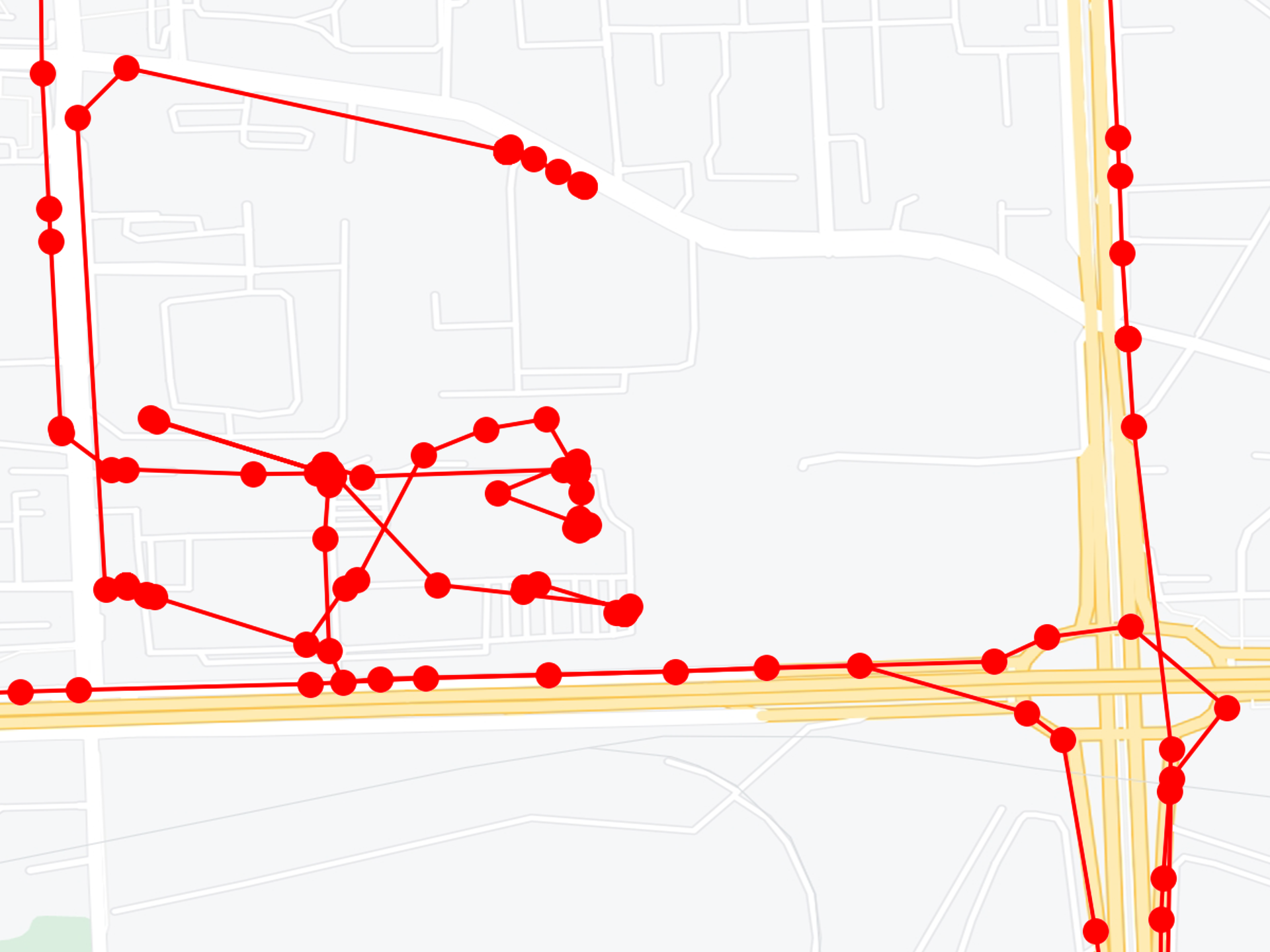}
    \subcaption{Sparse trajectory}
    \label{fig:case_sparse}
\end{subfigure}
    \begin{subfigure}[t]{0.49\linewidth}
    \centering
    \includegraphics[width=\linewidth]{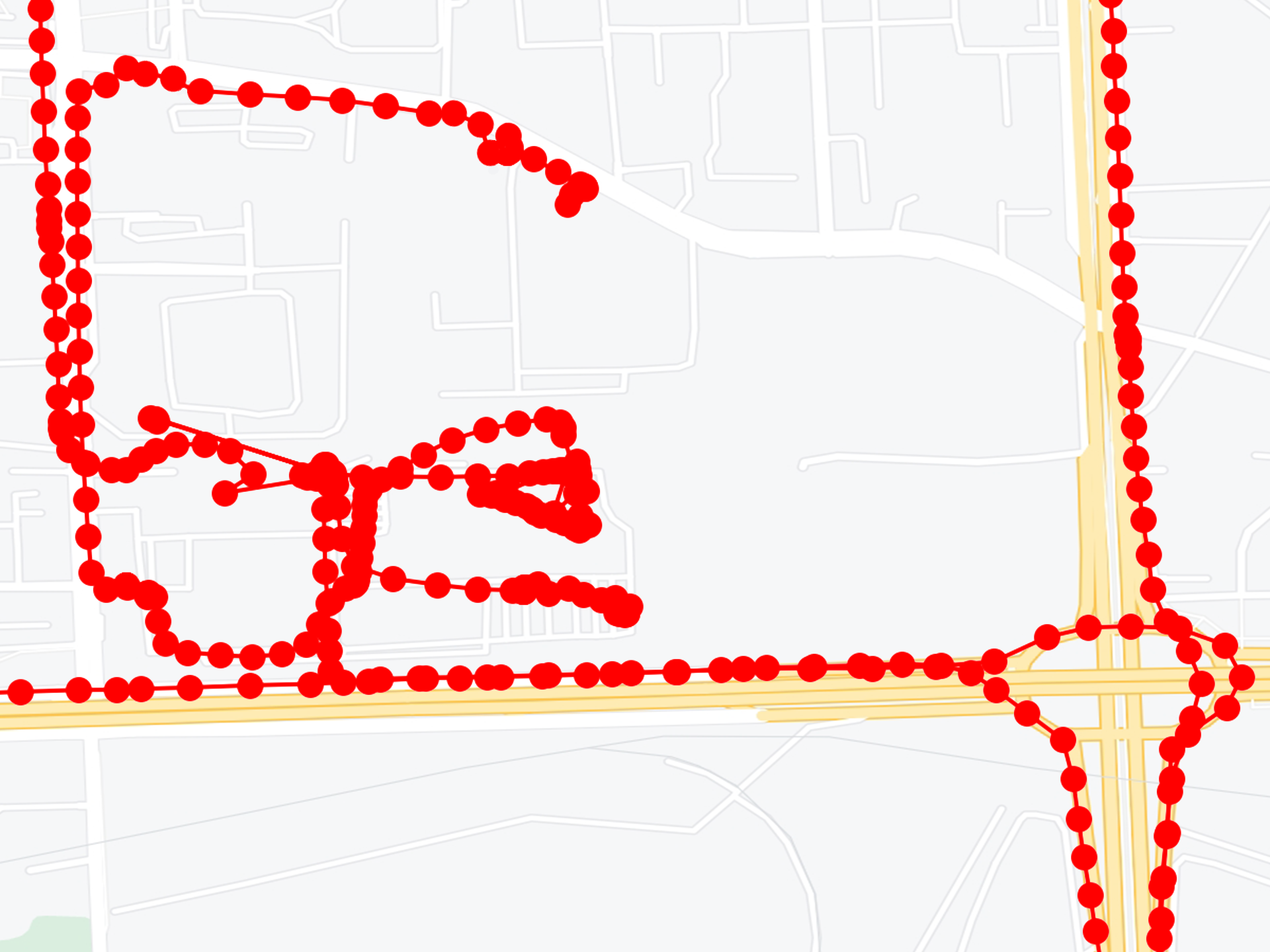}
    \subcaption{Recovered trajectory}
    \label{fig:case_recover_TW}
\end{subfigure}
 \caption{The dataset of: (a) Dense trajectories as ground truth. (b) Trajectories with the same sparsity as usual logistics scenario. (c) Recovered trajectories using \methodname{}.}
 \Description{The trajectories recovered by the proposed method makes sparse trajectory usable, and looks similar to ground truth.}
 \label{fig:case_study_visual}
\end{figure}

%% file: 5_RelatedWorks.tex
\section{Related Work}
\paragraph{Trajectory Recovery.}
Numerous efforts have been made to increase the accuracy of trajectory recovery. Certain map-based methods~\cite{HumanTrajCompletionTrans,TaxiTrajRec,MTrajRec,RNTrajRec} utilize a predefined set of points of interest (POI) or a pre-collected road network as strong prior knowledge. In contrast, DHTR~\cite{TrajRecCalibKF} performs free-space trajectory recovery using RNN, which eliminates the need for road network information. However, for very long and sparse trajectories, RNNs struggle to capture dependencies among points. TrajBERT~\cite{TrajBert} and AttnMove~\cite{AttnMove} are methods based on Transformers~\cite{Transformer}, which excel in capturing long-term dependencies within sequence.

\paragraph{Diffusion-based Trajectory Generation/Recovery.}

Existing approaches have successfully applied diffusion models to generate or recover time series data, including CSDI~\cite{CSDI} and SSSD~\cite{SSSD}. In particular, PriSTI~\cite{PriSTI} is a diffusion-based method designed for time series recovery, using spatio-temporal conditions and geographic factors for more accurate recovery. Moreover, another thread of existing work uses diffusion models for trajectory generation rather than recovery. For example, DiffTraj~\cite{DiffTraj} employs UNet~\cite{UNet} to perform denoising steps with several simple contexts such as trajectory length, starting and ending points. More recently, some diffusion-based methods utilize road networks to guide the generation process, such works include Diff-RNTraj~\cite{diff_rn_traj} and ControlTraj~\cite{ControlTraj}.

%% file: 6_Conclusion.tex
\section{Conclusion}

In conclusion, this paper presents \methodname{}, a diffusion-based framework for robust trajectory recovery. At its core, \methodname{} incorporates the State Propagation Diffusion Model (SPDM), which introduces a novel state propagation pipeline that enables knowledge sharing across denoising steps. This design effectively addresses the spatio-temporal irregularities of real-world trajectory data, leading to more accurate reconstruction of sparse segments while reducing redundant computation. Extensive experiments on multiple real-world datasets confirm the superior robustness and efficiency of \methodname{} over state-of-the-art methods. By providing precise and efficient trajectory recovery, our work enhances the data reliability for large-scale location-based web services and smart city applications.